\documentclass[letterpaper, 10 pt, conference]{ieeeconf}  % Comment this line out if you need a4paper

\IEEEoverridecommandlockouts                              % This command is only needed if 
\usepackage{amsmath,graphicx,amsfonts}
\usepackage{cite}

\title{\LARGE \bf
Unified Cross-Modal Attention-Mixer Based Structural-Functional Connectomics Fusion for Neuropsychiatric Disorder Diagnosis}

\author{Badhan Mazumder$^{1}$, Lei Wu$^{2}$, Vince D. Calhoun$^{2}$ and Dong Hye Ye$^{1,2, \dag}$
\thanks{$^{1}$Department of Computer Science, Georgia State University, Atlanta, GA, USA.}%
\thanks{$^{2}$Tri-Institutional Center for Translational Research in Neuroimaging and Data Science (TReNDS), Georgia State University, Georgia Institute of Technology, and Emory University, Atlanta, GA, USA.}%
\thanks{$^{\dag}$Corresponding Author: \texttt{dongye@gsu.edu}}
}

\begin{document}

\maketitle
\thispagestyle{empty}
\pagestyle{empty}

%%%%%%%%%%%%%%%%%%%%%%%%%%%%%%%%%%%%%%%%%%%%%%%%%%%%%%%%%%%%%%%%%%%%
\begin{abstract}
Gaining insights into the structural and functional mechanisms of the brain has been a longstanding focus in neuroscience research, particularly in the context of understanding and treating neuropsychiatric disorders such as Schizophrenia (SZ). Nevertheless, most of the traditional multimodal deep learning approaches fail to fully leverage the complementary characteristics of structural and functional connectomics data to enhance diagnostic performance. To address this issue, we proposed ConneX, a multimodal fusion method that integrates cross-attention mechanism and multilayer perceptron (MLP)-Mixer for refined feature fusion. Modality-specific backbone graph neural networks (GNNs) were firstly employed to obtain feature representation for each modality. A unified cross-modal attention network was then introduced to fuse these embeddings by capturing intra- and inter-modal interactions, while MLP-Mixer layers refined global and local features, leveraging higher-order dependencies for end-to-end classification with a multi-head joint loss. Extensive evaluations demonstrated improved performance on two distinct clinical datasets, highlighting the robustness of our proposed framework.
\newline
\indent \textit{Clinical relevance}— 
This study presents a novel approach to effectively integrate structural and functional brain connectivity for improved understanding and diagnosis of neuropsychiatric disorder.
% This is a brief additional statement on why a this might be of interest to practicing clinicians. Example: This establishes the anesthetic efficacy of 10\% intraosseous injections with epinephrine to positively influence cardiovascular function.
\end{abstract}

\section{INTRODUCTION}

The sophisticated neurobiological mechanism of human brain consists of neurons, circuits, and subsystems that interact in mysterious manners. Schizophrenia (SZ) considered as one of the severe mental disorders, usually manifests in late adolescence and is characterized by symptoms such as delusions, hallucinations, and paranoia \cite{I_1}. Despite several decades of research, its pathogenesis continues to remain unexplained \cite{I_5}.  Previous clinical research \cite{I_2,I_3,I_4,I_9} underscored the significance of investigating both structural and functional brain connectivity to gain a deeper understanding of SZ, implying that combining both modalities \cite{I_6,I_7,I_8,I_10,I_LR_9} would lead to better diagnostic accuracy. 

Brain networks, also known as the connectome, can be considered as complex graphs in which anatomical regions are termed as nodes and the connectivities underlying them are represented as links. To leverage this, graph neural networks (GNNs) have become increasingly popular in recent years for the analysis of both structural and functional connectomics data \cite{I_LR_3,I_LR_4,I_LR_5,I_LR_6,I_LR_7,I_LR_8,I_LR_10,I_LR_12,I_LR_13}. In contrast to shallow models \cite{I_LR_1,I_LR_2,I_LR_11}, GNNs hold better representation capabilities by converting adjacency matrices and node characteristics into a lower-dimensional space, which makes them perfect for studying nonlinear connectomes. 
\begin{figure*}[htb]
% \begin{minipage}[b]{1.0\linewidth}
  \centering
  \centerline{\includegraphics[width=\linewidth,height=7 cm]{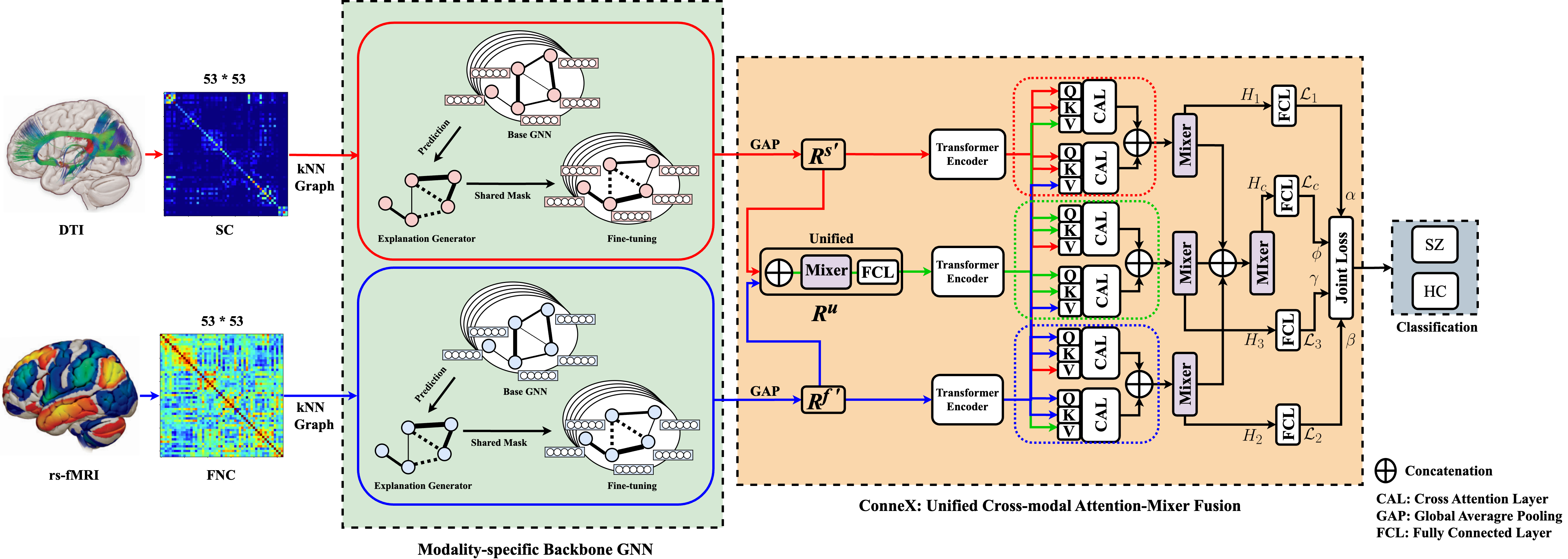}}
  \vspace{-5pt}
% \end{minipage}
\caption{Method overview: GNNs were trained on modality-wise connectome graphs, followed by an explanation generator producing a shared mask across subjects, which was utilized for fine-tuning the base GNNs to enhance learned representations. These structural-functional representations were then cross-attended using our proposed fusion method ConneX where these representations were combined and fed as an additional unified representation (highlighted with green arrows) followed by a multi-head joint loss for final classification task.}
  \vspace{-10pt}

\label{fig:fig1}
\end{figure*}
However, most of the existing GNN based approaches \cite{I_LR_3,I_LR_4,I_LR_5,I_LR_8,I_LR_10} overlook disorder-specific explanations to enhance GNN representations in a multi-view scenario, despite the fact that graph-level connectome-based disorder analysis \cite{I_LR_6} confirms individuals with the same disorder demonstrate similar connectome patterns. Recently, attention-based methods \cite{I_LR_8} have been widely used to capture inter-modal associations in connectome modalities, but they struggle to capture intra-modal interactions and fail to explicitly address local and global dependencies. Furthermore, the issue of modality predominance \cite{I_11}—when the training algorithm favors one modality over the others—occurs very often in these contemporary multimodal deep models \cite{I_LR_3,I_LR_4,I_LR_5,I_LR_10,I_LR_14,I_LR_15}.

To address the aforementioned limitations, we proposed a framework to learn explainability enhanced modality-wise connectomics representation and perform effective multimodal fusion to improve diagnostic accuracy for classification of SZ individuals from healthy control (HC). To incorporate disorder-specific insights, GNNs were applied to both connectomes separately, followed by an explanation generator producing modality-specific masks to highlight disorder-related connections, which were then utilized to enhance the connectome data and fine-tune the base GNNs for improved feature representation. For multimodal fusion, we proposed a unified cross-modal attention-mixer approach named ConneX that integrated a multimodal transformer architecture with multilayer perceptron (MLP)-Mixer \cite{M_6} layers introducing a third input branch representing the joint features of both connectome modalities. This additional branch allowed the model to access complementary contextual information that may not be fully captured by cross-attention alone, thereby enhancing the understanding of intra- and inter-modal relationships between structural and functional connectome. On the other hand, the integrated MLP-Mixer layers effectively combined those multimodal features by facilitating both local and global interactions. Furthermore, to resolve the optimization imbalance problem, we employed a multi-head joint loss function by adding simple loss weighting mechanism, which improved the downstream classification task's performance in an end-to-end style.

To sum up, our key contributions are listed as follows:
\begin{itemize}

\item We enhanced modality-specific connectome representations by incorporating disorder-specific insights using a multi-view explainable GNN approach.
\item We proposed a novel multimodal fusion architecture employing unified modality representation to capture both intra- and inter-modal interactions within the connectome using cross-attention, while also exploiting global and local dependencies across modalities through integrated MLP-Mixer layers.
\item We employed a multi-head joint loss function in our multimodal approach to address the issue of modality predominance, ensuring more balanced multimodal fusion and enhancing classification performance.
\item Comprehensive experiments on two clinical datasets (for SZ: FBIRN and COBRE) confirmed that our proposed framework can surpass state-of-the-art (SOTA) approaches and relevant fusion baselines.
\end{itemize}

\vspace{-1pt}
\section{METHODOLOGY}
Let's consider the provided dataset as $ C= \left \{ \left ( D_{n}^{s}, D_{n}^{f} \right ), W_{n} \right \}_{n=1}^{P} $ with a total of $P$ subjects where for each subject $n$,  $D^{s}$ and $D^{f}$  denotes the structural and functional connectomics modality, respectively, while $W$ presents the associated label ($0$: HC or $1$: SZ). As illustrated in Fig. \ref{fig:fig1}, our objective is to obtain a low-dimensional fused embedding for each \(n_{th}\) subject across both connectomics modalities to perform improved graph-level classification task. \textit{Note:} Throughout the following sections, the superscript $^{*}$ denotes either modality to simplify notation.
 
\subsection{Modality-specific Backbone GNN}
\subsubsection{Connectome Graph Formulation}
To define each connectome modality, weighted graph $ D^{*} = \left( F, G^{*}, E^{*} \right) $ was formed where $ F = \{f_i\}_{i=1}^{M} $ denotes nodes with size $M$ corresponding to the brain networks in connectome, and $ G^{*} = F \times F $ presents edges (connections between brain networks) defined by weighted adjacency matrix $E^{*}\in \mathbb{R}^{M \times M}$. 
% From $ E^{*} $, a k-NN graph was derived by selecting the top $ k $ neighbors for each node, constrained by the $ k $ value to adjust graph sparsity which was proven to improve diagnostic performance by discarding irrelevant connections 
From $E^{*}$, a k-NN graph was derived by selecting the top $k$ neighbors for each node, with $k$ controlling graph sparsity—an approach proven to enhance diagnostic performance by discarding irrelevant connections \cite{I_LR_10}. We investigated the effect of different $ k $-values ($3, 5, 10, 20 $) and used $ k = 5 $ as it yielded the best outcomes. Local degree profiles (LDP) \cite{I_LR_3} were determined from the formulated graphs and employed as node feature $l_{q} = [Deg(q)\oplus Mean(U_{q})\oplus Std(U_{q}) \oplus Min(U_{q}) \oplus Max(U_{q})] $. Here, $\oplus $ represents concatenation and $ U_{q} = \{ Deg(m) \mid (q, m) \in G^{*} \} $ denotes degree statistics of node $ q $ within two-hop neighborhood.
\subsubsection{Explainabiliy Enhanced GNN Representations}
After constructing modality-wise connectome graphs, we treated each modality as separate views and fed them into the Residual Gated Graph Convolutional network (RGGCN) \cite{M_3}. Compared to conventional Graph Convolutional Networks (GCNs) \cite{M_1}, RGGCNs have demonstrated superior performance in graph classification tasks, particularly when dealing with complex and variable-sized graph-structured data, such as functional and structural connectomes \cite{M_2, M_3}. We employed RGGCN followed by global average pooling (GAP) layer to obtain graph-level feature representations, denoted as $ R^{s} $ and $ R^{f} $, for the structural and functional modalities, respectively, and to generate the corresponding predictions, $\hat{W}^{*}$.  

Subsequently, to derive a globally shared explanation graph for each modality, we integrated a global explanation generator \cite{M_4,I_LR_6} to learn modality-specific globally shared edge mask $ Y^{*} \in \mathbb{R}^{M \times M} $. These masks were consistently applied across the brain networks of all subjects within the dataset. Specifically, our objective was to maximize the degree of agreement between the predictions $\hat{W}^{*}$ obtained from the original graph $ D^{*}$ and $\hat{W}^{*'}$ from the explanation-based graph $ D^{*'}=\left ( F, G^{*}, E^{*'} \right)$, which was generated by applying the learned mask $Y^{*}$ where $ E^{*'} = E^{*}\odot \sigma (Y^{*}) $. Here, $\sigma$ and $\odot$ represents the sigmoid function and element-wise multiplication, respectively.  

We leveraged the generated disorder-relevant connections by fine-tuning our base GNNs to learn more enhanced feature representations for the both structural and functional modalities, denoted as $ R^{s'}$ and $ R^{f'}$, respectively. Overall, the training process and explanation generator were integrated into a unified framework for learning GNN representations and explanations in a closed-loop fashion.
\vspace{-2pt}
\subsection{ConneX for Structural-Functional Fusion}
% \subsubsection{Capturing Intra- and Inter-Modal Interactions}
% \subsubsection{Leveraging Global and Local Dependencies}
% \subsubsection{Multi-head Joint Loss Classification}
We designed ConneX to address the high complexity nature of connectomics data. A cross-attention-based architecture was developed by introducing an additional unified input branch and integrating MLP-Mixer layers at the early and later stages of the fusion block to achieve impactful representations across modalities.
\subsubsection{Unified Representation Generation}
To generate the unified feature representation $ R^{u} $, the obtained modality-specific GNN representations $ R^{s'}$ and $ R^{f'}$ were firstly concatenated and then passed through one MLP-Mixer \cite{M_6} layer before feeding it to a full connected layer (FCL) for dimensionality reduction purpose. 

The MLP-Mixer was used here to leverage both global (subject-level) and local (feature-level) interactions within the fused representation, allowing for more refined feature fusion compared to direct concatenation, which may fail to capture these intricate dependencies. The token mixing block (Equ. \ref{E2}) in MLP-Mixer enhanced global relationships across subjects by allowing information to flow effectively across them and capturing higher-level dependencies and population-wise disorder patterns. Subsequently, the channel mixing block (Equ. \ref{E3}) in MLP-Mixer operated at the local level with feature vector per subject refining the internal representation by improving interactions among the fused structural and functional features, highlighting the most significant disorder patterns while mitigating noise. 
\begin{equation}
 A=Z^{T}+X_{2}\delta [X_{1}LayerNorm(Z^T)]
\label{E2}
\end{equation}
\begin{equation}
B=A^{T}+X_{4}\delta [X_{3}LayerNorm(A^T)]
\label{E3}
\end{equation}

Here, $Z$ defines the input for MLP-Mixer layer, $(\cdot)^T$ indicates the transpose operation, $LayerNorm(\cdot)$ presents the layer normalization, $\delta$ denotes GELU activation function. $X_{1}, X_{2}, X_{3}$ and $X_{4}$ represents the trainable weight matrices associated with the respective layers.

\subsubsection{Cross-modal Attention-Mixer Fusion}
Each modality representation was then fed into its dedicated encoder, which employed a multi-head self-attention mechanism \cite{M_5} using Equ.\ref{E1} followed by a FCL. Both layers included residual connections and layer normalization. 
\begin{equation}
 Attention(Q,K,V)= softmax(QK^{T}/ \sqrt{d_{k}})V
\label{E1}
\end{equation}
Conventional cross-attention \cite{M_5} fusion primarily focuses on capturing correlations between different modalities. In contrast, our proposed method not only models these intermodal interactions but also incorporates a unified representation ($ R^{u} $) to be input into the multimodal transformer. This additional joint representation enriches the model's contextual understanding, enabling it to better capture intricate relationships across connectome modalities while minimizing the impact of noise or irrelevant features. As a result, the model demonstrates improved performance and greater robustness when processing diverse input sequences.

The self-attention layers (Equ.\ref{E1}) allowed each encoder to focus on key features within its own modality (intra-modal interaction), enabling the model to extract relevant information from each individual source. Afterward, the embeddings from each encoder were merged via six cross-modal attention layers, where the value matrix ($V$) was shared with the key ($K$) and query ($Q$) matrices of the other modalities. This shared matrix structure enables both inter-modal interaction—capturing the mutual influence between structural and functional connectomes—and intra-modal interaction, refining relationships within each modality. 
% \subsubsection{Feature Refinement via MLP-mixer}

For leveraging both global and local interactions within the six cross-modal attention layers outputs, we employed the MLP-Mixer architecture in a multi-view setting to capture higher-level dependencies and population-wise disorder patterns that may not be fully addressed by cross-attention module alone. Given that we had three different input modalities: structural ($ R^{s'}$), functional ($ R^{f'}$), and the unified representation ($ R^{u}$); we structured the outputs of these cross-modal attention layers based on three distinct views. As illustrated in Fig. \ref{fig:fig1}, the outputs of the cross-attention layers, where the $Q$ and $K$ matrices were derived from the same modality, were concatenated and fed into a seperate MLP-Mixer layer. Subsequently, the outputs of the multi-view MLP-Mixer layers: $H_{1}, H_{2}, H_{3}$ were concatenated and passed through another MLP-Mixer layer to obtain the final representation $H_{c}$ .
\begin{table*}[t]
\caption{Comparison with SOTA approaches with 5-fold cross-validation [Unit: \%] (Mean $\pm$ Standard Deviation).}
\begin{center}
\begin{tabular}{|c|c|c|c|c|c|c|} 
\hline
 Method  & \multicolumn{3}{c|}{FBIRN} & \multicolumn{3}{c|}{COBRE} \\  
\cline{2-7}
 & Accuracy & Precision & F1-score & Accuracy & Precision & F1-score \\  \hline
  GCNN \cite{I_LR_5} & 69.24$\pm$3.32 & 70.44$\pm$3.68 & 68.88$\pm$3.63 & 72.72$\pm$2.97 & 73.12$\pm$3.93 & 72.10$\pm$3.26 \\\hline
 BrainNN \cite{I_LR_3} & 85.14$\pm$1.14 & 86.67$\pm$1.69 & 84.89$\pm$1.38 & 80.95$\pm$1.82 & 81.67$\pm$2.48 & 80.74$\pm$2.04 \\\hline
 Joint DCCA \cite{I_LR_10} & 83.16$\pm$1.18 & 85.78$\pm$1.85 & 83.67$\pm$1.38 & 81.89$\pm$1.66 & 82.19$\pm$2.19 & 81.87$\pm$2.52 \\\hline
 \textbf{Proposed} & \textbf{88.53$\pm$1.49} & \textbf{93.33$\pm$1.33} &\textbf{ 85.56$\pm$1.29} & \textbf{85.71$\pm$1.11} & \textbf{86.67$\pm$1.05} & \textbf{84.05$\pm$1.25} \\
\hline

\end{tabular}
\label{Table_1}
\end{center}
\end{table*}

\begin{table*}[t]
\caption{Ablation experiments results with 5-fold cross-validation [Unit: \%] (Mean $\pm$ Standard Deviation; Y = Yes; N = No).}
\begin{center}
\begin{tabular}{|c|c|c|c|c|c|c|c|c|c|} 
\hline
Modality & Explanation  & Unified & Fusion  & \multicolumn{3}{c|}{FBIRN} & \multicolumn{3}{c|}{COBRE} \\  
\cline{5-10}
 &  Enhanced & Representation& Method & Accuracy & Precision & F1-score & Accuracy & Precision & F1-score \\ 
 % &  &  $R^{u}$ &  &  &  &  &  &  &  \\ 
  \hline

SC & N & - & - & 71.43$\pm$1.46 & 71.33$\pm$2.04 & 66.00$\pm$1.81 & 74.29$\pm$1.74 & 78.72$\pm$2.22 & 74.33$\pm$1.12 \\ \hline
SC & Y & - & - & 76.00$\pm$1.79 & 74.17$\pm$2.35 & 70.40$\pm$1.90 & 78.85$\pm$1.26 & 80.20$\pm$2.10 & 78.47$\pm$1.38 \\\hline
FNC & N & - & - & 73.33$\pm$1.94 & 75.17$\pm$2.35 & 71.93$\pm$2.06 & 76.19$\pm$1.36 & 79.54$\pm$1.72 & 75.24$\pm$1.51 \\\hline
FNC & Y & - & - & 77.14$\pm$1.67 & 77.33$\pm$2.13 & 75.87$\pm$1.42 & 80.67$\pm$1.15 & 81.67$\pm$2.47 & 77.57$\pm$1.42 \\\hline
SC+FNC & Y & N & Concat & 78.38$\pm$1.80 & 77.83$\pm$1.90 & 77.46$\pm$1.66 & 80.75$\pm$1.83 & 81.90$\pm$2.33 & 77.71$\pm$1.10 \\ \hline
SC+FNC & Y & N & DCCA & 80.26$\pm$1.65 & 78.48$\pm$2.34 & 79.13$\pm$1.75 & 81.33$\pm$1.21 & 82.71$\pm$1.31 & 80.68$\pm$1.29 \\ \hline
SC+FNC & Y & N & Cross-Att & 84.55$\pm$1.26 & 82.43$\pm$1.65 & 80.48$\pm$1.71 & 82.16$\pm$1.18 & 84.71$\pm$1.74 & 80.45$\pm$1.48 \\\hline
SC+FNC & Y & N & ConneX & 84.67$\pm$1.67 & 84.83$\pm$1.94 & 81.35$\pm$1.54 & 82.84$\pm$1.81 & 84.77$\pm$1.76 & 81.86$\pm$1.91 \\\hline
SC+FNC & Y & Y & Concat & 80.55$\pm$1.58 & 80.36$\pm$1.22 & 79.68$\pm$1.11 & 80.88$\pm$0.97 & 83.18$\pm$2.15 & 81.39$\pm$1.16 \\\hline
SC+FNC & Y & Y & DCCA & 83.92$\pm$1.13 & 84.58$\pm$2.05 & 82.37$\pm$1.03 & 81.67$\pm$2.49 & 83.67$\pm$2.42 & 81.33$\pm$2.16 \\\hline
SC+FNC & Y & Y & Cross-Att & 85.34$\pm$1.18 & 86.39$\pm$1.11 & 84.57$\pm$1.13 & 83.01$\pm$1.35 & 84.91$\pm$1.74 & 82.41$\pm$1.23 \\\hline
SC+FNC & \textbf{Y} & \textbf{Y} & \textbf{ConneX} & \textbf{88.53$\pm$1.49} & \textbf{93.33$\pm$1.33} & \textbf{85.56$\pm$1.29} & \textbf{85.71$\pm$1.11} & \textbf{86.67$\pm$1.05} & \textbf{84.05$\pm$1.25} \\
 %&  &  & \textbf{[Proposed]} &  &  &  &  &  &  \\
\hline

\end{tabular}
\label{Table_2}
\end{center}
\end{table*}

\vspace{-5pt}
\subsection{Multi-head Joint Loss Classification}
In order to achieve balanced multimodal fusion \cite{I_11} and improved classification performance, classification heads were incorporated to each of the multi-view MLP-Mixer outputs, as well as to the final fused feature representation. This approach resulted in a total of four separate classification branches as defined in Equ. \ref{E4} below: 
\begin{equation}
\bar{W}_{j}= \underset{c}{argmax} \quad\sigma(FCL(H_{j})), \quad j\in {c,1,2,3}
\label{E4}
\end{equation}
here, $\sigma$ denotes the sigmoid function and $\bar{W}_j$ represents the predicted output from the classification branch indexed by $j \in \{c,1,2,3\}$, where $1,2,3$ correspond to modality-specific views and $c$ corresponds to the final fused feature representation.

% where σ is the Sigmoid function in the case of binary classification

The network was optimized by jointly minimizing the multi-head joint loss, which determined the contribution of each branch to the overall optimization and forced our fusion network to learn optimal representations through a unified objective function as given below in Equ. \ref{E5}.
\begin{equation}
\begin{aligned}
\mathcal{L}(\bar{W}_{1},\bar{W}_{2},\bar{W}_{3},\bar{W}_{c},W) &= 
\alpha \mathcal{L}_{1}(\bar{W}_{1},W) + 
\beta  \mathcal{L}_{2}(\bar{W}_{2},W) \\
&\quad + \gamma  \mathcal{L}_{3}(\bar{W}_{3},W) 
+ \phi   \mathcal{L}_{c}(\bar{W}_{c},W)
\end{aligned}
\label{E5}
\end{equation}

Here, the weighting factors $\alpha$, $\beta$, $\gamma$, and $\phi$ sum to 1 and determine the extent to which each branch contributes to the overall optimization process.

\section{RESULTS AND DISCUSSIONS}
\subsection{Dataset and Data Pre-processing}
We utilized subsets from the Function Biomedical Informatics Research Network (FBIRN) \cite{R_1} and the Center for Biomedical Research Excellence (COBRE) \cite{R_2} datasets to evaluate the validity of our framework. These datasets include diffusion tensor imaging (DTI) and resting-state functional magnetic resonance imaging (rs-fMRI) scans from HC and individuals with SZ. The FBIRN subset consisted of 165 participants aged 18 to 62 years, including 93 individuals with SZ and 72 HC. The COBRE subset comprised a total of 152 participants, including 64 individuals with SZ and 88 classified as HC, with age ranging from 18 to 65 years. These subsets were selected based on the availability of both imaging modalities.

To generate subject-specific 53 $\times$ 53 functional network connectivity (FNC) matrices from rs-fMRI data, we employed the \textit{Neuromark} pipeline \cite{R_3}, an automated spatially constrained method involving independent component analysis (ICA). At first, for every subject, distinct functional components and their corresponding time courses were determined via an adaptive-ICA technique.  Then we measured the correlation among the temporal patterns of 53 intrinsic connection networks (ICNs) to obtain the subject-wise FNC matrices. Structural connectivity (SC) was obtained from DTI by estimating diffusion tensors via FSL \cite{R_6} and conducting whole-brain deterministic tractography by CAMINO \cite{R_4}. The \textit{Neuromark} atlas \cite{R_3} was then spatially aligned to match the native space, and the fractional anisotropy map was warped into MNI space. Lastly, streamlines traversing each pair of atlas networks were identified and quantified.

\subsection{Implementation Details}

Our proposed framework was developed in PyTorch and run on an NVIDIA V100 GPU. The experiments were conducted using 5-fold cross-validation with an 80:20 training-to-testing split ratio. The backbone GNNs consisted of 5 layers, each with a channel size of 32. For the FBIRN dataset, training was performed for 1000 epochs using the Adam optimizer with a learning rate of 0.001, a batch size of 16, and a dropout rate of 0.6, resulting in deep feature vectors of dimension 32. For the COBRE dataset, all parameters remained the same, except the training was conducted for 1200 epochs.

During the modality fusion step, the GNN backbones were kept frozen, and a grid search was performed to determine the optimal hyperparameters for our fusion network. Accordingly, for both FBIRN and COBRE, the fusion block was trained for 300 epochs using the Adam optimizer with a learning rate of 0.0001, with the batch size of 8 for FBIRN and 4 for COBRE.

\subsection{Quantitative Evaluation}
\subsubsection{Comparison with SOTA approaches}
Table \ref{Table_1} represents a comparison between the classification outcomes of our proposed method and three other state-of-the-art (SOTA) approaches, highlighting that our proposed approach achieves superior performance compared to methods employing different multimodal fusion strategies for structural-functional connectomics. These findings indicate that, in comparison to contrastive learning \cite{I_LR_5}, canonical correlation analysis objective \cite{I_LR_10} and GNN convolution based \cite{I_LR_3} connectome fusion methods, our cross-modal attention based end-to-end simultaneous fusion approach produced a more robust fused representation by leveraging both intra-inter modal and global-local interaction effectively, leading to improved classification performance.

\subsubsection{Ablation Experiments}

To validate the effectiveness of each module within our proposed framework, we conducted detailed ablation experiments as outlined in Table 2. First, we validated the impact of using explainability enhancement in our GNN backbones. For confirming the significance of including disorder-specific connections, we examined the effect of using both fine-tuned GNN-based feature embeddings and only base GNN embeddings. Fine-tuning the GNN backbones with disorder-specific explanations increased classification accuracy for both the structural (FBIRN: from 71.43\% to 76.00\%; COBRE: from 74.29\% to 78.85\%) and functional connectomes (FBIRN: from 73.33\% to 77.14\%; COBRE: from 76.19\% to 80.67\%), confirming its effectiveness. Additionally, it indicates that, compared to using SC and FNC separately in a unimodal setup, multimodal approaches improve overall diagnostic performance, highlighting the advantage of integrating both connectomics for disorder analysis.

The effectiveness of our proposed ConneX was evaluated by performing comparative experiments across modalities with three different fusion methods: simple concatenation (Concat), deep canonical correlation analysis (DCCA) \cite{M_7}, and vanilla cross-attention (Cross-Att) \cite{M_5}. As we incorporated a unified structural-functional representation ($R^{u}$) as an additional input branch in our proposed fusion approach, we also examined its impact by evaluating the model's performance without it. In DCCA fusion, we employed two DCCA modules in case of with $ R^{u} $ and one module without $ R^{u} $. For Cross-Att fusion, we stacked the outcomes of the cross-attention layers (with $ R^{u} $: 6 layers, without $ R^{u} $: 2 layers). In all three baseline fusion methods, we fed the final fused embeddings into an FCL for classification. Table \ref{Table_2} demonstrates the superiority of our proposed fusion network over these baseline fusion methods and highlights the significance of including the unified representation as an additional input modality. It reports that the inclusion of the unified representation improved the classification performance for all fusion methods, particularly for ConneX up to 3.86\% for FBRIN and 2.87\% for COBRE in terms of accuracy which directly validate our hypothesis to introduce the unified representation to perform effective fusion.

\subsection{Qualitative Evaluation}

\begin{figure}[htb]
% \begin{minipage}[b]{1.0\linewidth}
  \centering
  \centerline{\includegraphics[width=\linewidth,height=5.5 cm]{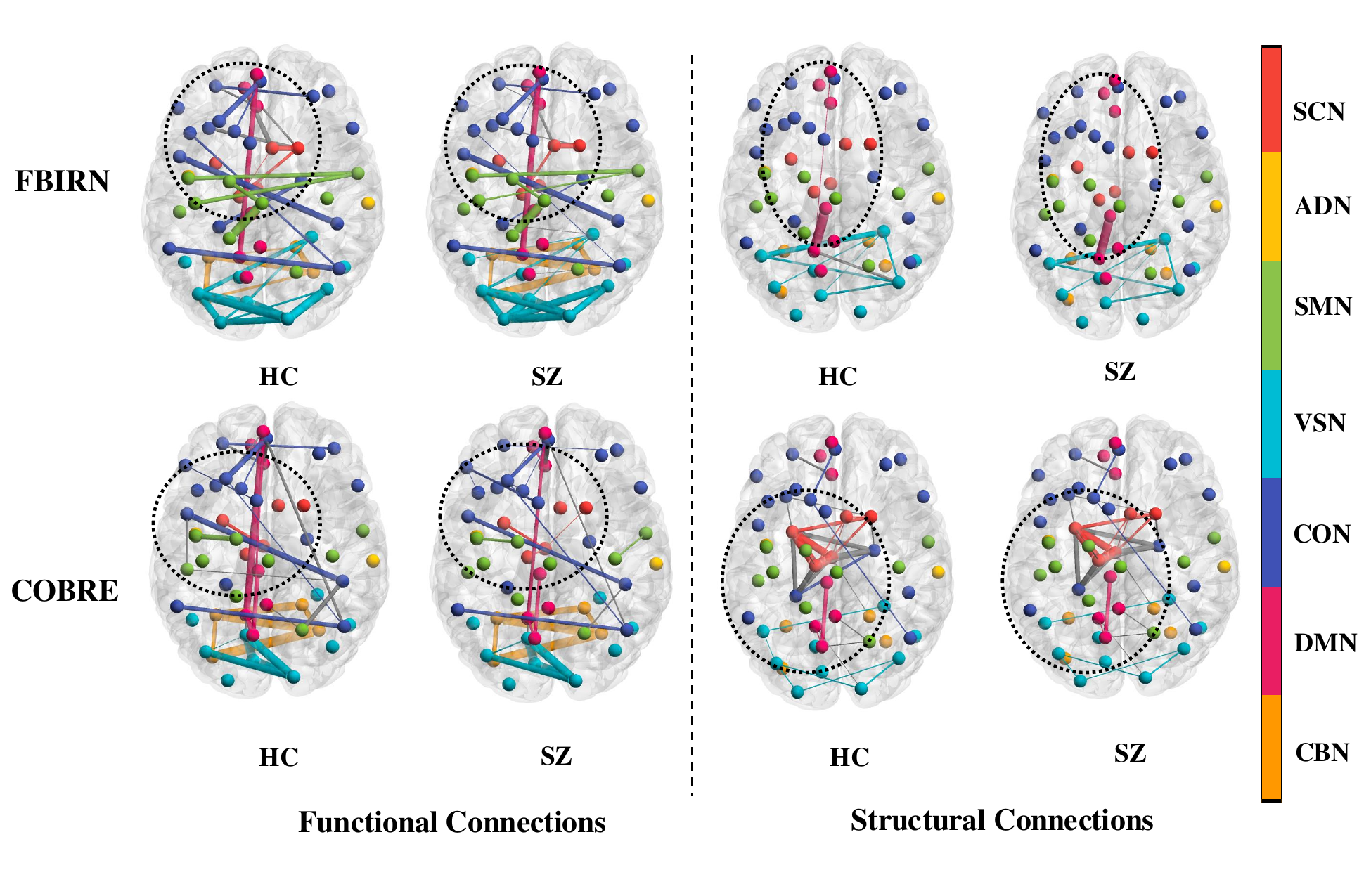}}
  \vspace{-5pt}
% \end{minipage}
\caption{Axial visualization of the top 100 most significant brain network connections, spanning the subcortical (SCN), auditory (ADN), sensorimotor (SMN), visual (VSN), cognitive control (CON), default mode (DMN), and cerebellar (CBN) networks, for both structural and functional connectomes across SZ and HC groups in the FBIRN and COBRE datasets. Connections within the same brain network  are highlighted with distinct colors, while inter-network connections are represented in gray. The edge width reflects the weight in the explanatory graph.}
\vspace{-5pt}
\label{fig:fig2}
\end{figure}

Explanation of significant connections were provided by the learned globally shared explanation mask $Y^{*}$. From the obtained mask $Y^{*}$, we generated explanation graph for both structural and functional connectomics separately. Firstly, $Y^{*}$ was applied on each graph $D^{*}$ for both modalities and after that, for both SZ and HC group, we averaged and normalized them all together and demonstrated top 100 group-specific significant connections. The illustrations in Fig. \ref{fig:fig2} depicts the top 100 important connections for both structural and functional modalities, aiding in the identification of individuals with SZ.

Fig. \ref{fig:fig2} depicts that denser connectivity was observed in both structural and functional connectomes for HC compared to individuals with SZ across both datasets. In the functional connectome, for the FBIRN dataset, in comparison to HC, SZ individuals displayed significant less interactions within the Default Mode Network (DMN) and Cognitive Control Network (CON), while in the COBRE dataset, less interactions were found within the DMN alone. For the structural connectome, SZ individuals in FBIRN exhibited fewer interactions within the DMN, CON, and Sensorimotor Network (SMN). In the COBRE dataset, similar interactions were observed within the SMN, Subcortical Network (SCN), and Visual Network (VSN). These obtained outcomes confirm the clinical observation of SZ disrupting both structural and functional dynamical states of the human brain \cite{I_6,I_7,I_8,I_10,I_LR_9} and also are in alignment with earlier research studies \cite{I_LR_2,I_LR_1,I_LR_10,I_LR_9}.

\vspace{-5pt}
\section{CONCLUSIONS}
In this paper, we introduced ConneX, a novel multimodal connectomics fusion approach for improved  classification of individuals with SZ. ConneX was designed to effectively capture both intra- and inter-modal interactions while modeling dependencies at both global and local feature levels. At the core of our approach is the integration of a unified cross-modal attention-mixer fusion network with an explainability enhanced backbone GNN block, leading to improved performance for SZ diagnosis. Obtained experimental outcomes on the FBIRN and COBRE datasets demonstrated that our proposed framework outperforms all relative baseline methods, achieving robust and superior classification performance. For future work, we plan to extend ConneX to explore the temporal dynamics of multimodal connectomics and investigate its applicability to other neuropsychiatric disorders.

\bibliographystyle{IEEEtran}
\bibliography{refs}

\end{document}